\documentclass[12pt]{article}%
\usepackage{amsfonts}
\usepackage{amssymb}
\usepackage{graphicx}
\usepackage{setspace}
\usepackage[round]{natbib}
\usepackage{amsmath}%
\usepackage{amsthm}%
\usepackage{times} 
\usepackage{paralist}
\usepackage{subfigure}
\usepackage{xcolor}
\usepackage{multirow}
\usepackage{booktabs}
\usepackage{dsfont}
\usepackage{indentfirst}
\usepackage{enumerate}
\usepackage{longtable}
\usepackage{caption}
\usepackage{algorithm}

\usepackage{algpseudocode}%
\usepackage[hyperindex,breaklinks]{hyperref}
\hypersetup{colorlinks=true,       
   linkcolor=red,       
    citecolor=blue,        
    filecolor=magenta,      
    urlcolor=cyan           
}  

\usepackage{dcolumn}
\newcolumntype{L}{D{.}{.}{2,5}}

\usepackage[margin=1in]{geometry}

\title{Counterfactual Analysis via Large Language Models}

\author{Zonghao Yang \\
Stevens Institute of Technology \\
\texttt{zyang99@stevens.edu}}
\date{June 14, 2024}

\begin{document}
	
\maketitle
\vspace{-1cm}

\begin{abstract}

Counterfactual analysis aims to predict potential outcomes under hypothetical scenarios, offering valuable insights for decision-making. This paper investigates the application of large language models (LLMs), specifically the \textit{GPT-3.5} model, for counterfactual analysis. We focus on the online lending context, where the counterfactual return on investment (ROI) is crucial for evaluating different interest rate schemes. We begin by assessing the predictive performance of GPT and comparing it with advanced machine learning algorithms. The results show that prompt engineering can significantly enhance GPT's predictions, with the R-squared increasing from 1.97\% to 2.84\%, closely approaching the 3.48\% achieved by gradient-boosted regression. Subsequently, we utilize GPT to generate counterfactual ROIs under a set of alternative interest rates. GPT exhibits logical coherence and causal reasoning in its responses. The findings underscore the potential of LLMs as effective tools for counterfactual analysis in online lending, suggesting broader applications for LLMs in various predictive and decision-making contexts.

	
\end{abstract}


\newpage
\section{Introduction}

How do class sizes influence students' scholastic achievement \citep{angrist1999using}? How does the open rate of an email campaign change for an individual if the subject line is personalized instead of generic \citep{sahni2016personalization}? What is the recovery time for a patient who undergoes laparoscopic surgery instead of open surgery \citep{guller2004laparoscopic}? These questions hinge on understanding outcomes under hypothetical scenarios, an approach known as counterfactual analysis. Counterfactual analysis is a powerful tool for predicting the potential outcomes of various actions, uncovering causal relationships, and informing decision-making.

In this paper, we explore the use of large language models (LLMs) for counterfactual analysis. LLMs are gigantic neural networks trained on vast amounts of text data from diverse sources, including books, newspapers, and web pages. Because LLMs are trained on a much broader information set compared to alternative models, they can potentially approximate real-world complexities more closely. Although designed to generate text sequences, LLMs have demonstrated capabilities in understanding context \citep{chang2024survey} and imitating human behavior \citep{aher2023using}. For these reasons, LLMs hold significant potential for producing counterfactuals.

The experimental context requires both task complexity and the richness of information to engage LLMs in deliberate information processing and cognitive reasoning \citep{amit2013role}. We focus our analysis on online lending loans, although the insights gained from using LLMs for loan counterfactuals can be more general. In this context, the interest rate is the most important decision variable for online lending platforms. While the platforms can observe the loan outcomes under the original interest rates, the outcomes under alternative rates (i.e., counterfactuals) are not observable. Constructing such counterfactual measures is challenging but crucial for evaluating different interest rate schemes \citep{johnson2023fintech}. The goal of this paper is to provide counterfactual loan outcomes, specifically the return on investment (ROI), using large language models.

We utilize a large dataset of online loans from LendingClub (LC), including the original LC interest rates, and a set of alternative interest rates proposed in \cite{gopal2024pricing} specific to this dataset. The \textit{GPT-3.5} model developed by OpenAI serves as our primary large language model.\footnote{Unless otherwise noted, the results presented in this paper are based on the \textit{gpt-3.5-turbo-0125} model developed by OpenAI. For large-scale language model inference, we utilize the \href{https://platform.openai.com/docs/api-reference/introduction}{OpenAI API}.} To construct loan counterfactuals, we describe the loan and borrower information, along with the observed repayment under the original interest rate, in text format as a prompt and ask GPT to predict the loan outcome.

To distill relevant pre-trained knowledge and elicit cognitive reasoning from GPT, we perform prompt engineering, including role-play (asking GPT to act as a borrower or an expert on credit risk assessment) and employing different prompt strategies (e.g., zero-shot, chain-of-thought, and tree-of-thought prompts). Human-machine collaboration in complex, information-rich scenarios has demonstrated better predictions and decision-making \citep{lu2024information}. Given that LLMs mimic human behaviors, we also incorporate predictions from advanced machine learning (ML) algorithms into the prompts to improve the accuracy of the counterfactuals.

Since counterfactuals are unobserved, it is infeasible to directly test their quality. We validate LLMs as a suitable tool for counterfactual construction in two ways. First, we examine the predictive performance of GPT on realized loan outcomes and compare it against machine learning algorithms, including gradient-boosted regression. A model suitable for producing counterfactuals should accurately capture the underlying data-generating process of loan outcomes and make precise predictions for actual loan outcomes. Therefore, strong predictive power for observed outcomes can be seen as a necessary, but not sufficient, condition for a model's suitability to generate counterfactuals. Second, we use GPT to produce loan counterfactuals and assess whether its responses are logically coherent. While the best model for prediction may not necessarily be the best model for constructing counterfactuals, capturing the causal relationship between the interest rate and loan outcome is the ultimate testament to valid counterfactual analysis.

We use R-squared to measure the predictive performance of ROI under the original interest rates. 
The results show that prompt engineering significantly improves the predictive performance of GPT, with the R-squared increasing from 1.97\% to 2.84\%. 
The predictive power of GPT is comparable to that of gradient-boosted regression, which has an R-squared of 3.48\% on the same sample. Since the prompts used to generate the ROI prediction include the ML prediction, we conduct the forecast encompassing test \citep{chong1986econometric} to examine if GPT merely repeats the ML prediction. The test results suggest that the GPT prediction is not encompassed by the ML prediction, indicating that GPT identifies extra information about the loan outcome beyond the ML algorithm. Furthermore, the GPT responses for both the prediction and counterfactual tasks appear to be logically coherent, taking the ML predictions into consideration and reasoning through how the loan and borrower characteristics lead to a prediction.

The result that GPT can predict actual loan outcomes, with performance comparable to advanced machine learning algorithms, suggests that the LLM captures important aspects of borrower behavior. The fact that the GPT prediction sometimes deviates from the ML prediction shows that GPT critically evaluates the ML predictions based on its pre-trained knowledge and revises them with justifiable reasons when necessary. Lastly, the responses from GPT demonstrate its understanding of the causal relationship between the interest rate and the outcome and its ability to string together text into causal reasoning. Together, the results show that large language models are a novel, well-suited, and capable approach for counterfactual analysis.

The remainder of the paper is organized as follows. Section \ref{sec:literature} reviews the related literature on LLM applications and counterfactual analysis. Section \ref{sec:data} describes the loan dataset. Section \ref{sec:cntfact} presents the methodology and the empirical results. Section \ref{sec:conclusion} concludes.

\section{Literature}\label{sec:literature}

In recent years, there has been a rapid increase in the applications of large language models (LLMs) across various fields, including information systems research \citep{susarla2023janus}, medicine \citep{thirunavukarasu2023large}, and education \citep{kasneci2023chatgpt}. These applications offer new perspectives on addressing existing problems. For instance, \citet{yang2024large} employ LLMs as optimizers, where the optimization task is described in natural language, to solve linear regression and traveling salesman problems. To the best of our knowledge, we are the first to apply LLMs for counterfactual analysis, demonstrating that it is an suitable approach for generating loan counterfactuals. Additionally, we illustrate effective prompt engineering techniques within the loan context, which are potentially transferable to counterfactual analysis in other contexts.

There are various other methods to implement counterfactual analysis, including matching (e.g., \citealp{yahav2016tree}), randomized controlled trials (e.g., \citealp{kleinberg2018human}), and survival analysis (e.g., \citealp{stepanova2002survival}). Specifically for the loan context, \citet{johnson2023fintech} propose a reduced-form approach to generate counterfactual ROIs. They first identify a subsample of efficiently priced loans, and then regress the realized ROI on the interest rate while controlling for loan and borrower characteristics. The counterfactual ROI given a new interest rate can be determined by the fitted regression function. \citet{gopal2024pricing} model loan repayment as a survival process using the mixture cure model in survival analysis. They fit the model using historical loan samples to predict when the loan will be charged off or fully paid under an alternative interest rate. They compute the ROI based on the predicted loan status and duration.

Compared with the above alternatives, LLMs have a much more extensive information set. The massive pre-training data equips LLMs with knowledge about the counterfactual context, the ability to relate prompts to this knowledge, and causal reasoning to generate sensible counterfactuals. Additionally, LLMs are much more flexible in handling both input and output data.

\section{Data}\label{sec:data}

\subsection{LendingClub Loans}

LendingClub is the world's largest online lending platform. We collected 894,013 LendingClub loans from 2013 to 2020. These are unsecured personal loans originating online from January 2013 to May 2017, each with a maturity of 36 months. The dataset covers 83 loan and borrower characteristics. Loan characteristics include \textit{loan amount}, \textit{interest rate}, and \textit{loan purpose}; borrower characteristics include \textit{gross income}, \textit{FICO score}, and \textit{debt-to-income ratio}. Additionally, loan outcomes are included in the dataset, such as whether and when the loan is fully paid or charged off and how much principal and interest repayments are made.

In our dataset, the average loan amount is \$12,647, with an average interest rate of 11.9\% and charged-off rate of 12.24\%. We measure the profitability of a loan using its return on investment, calculated based on the cumulative discounted payments (CDP) received by lenders. This is done by discounting each monthly payment by an annual discount rate of 2\% and summing them up as the CDP. The ROI is then computed as the return comparing the CDP to the initial loan principal. The average ROI is 4.50\% in our sample. We focus on ROI for counterfactual construction.

We subsample loans originating in 2013 as the training sample, totaling 94,605 loans, which are used to train the benchmark machine learning algorithms. The remaining loans, totaling 799,408 and originating between January 2014 and May 2017, are used as the test sample for evaluation and fine-tuning of the large language model.

\subsection{Interest Rates}

To generate hypothetical scenarios in the loan context, we adopt an alternative set of interest rates proposed by \cite{gopal2024pricing}. This set of interest rates is introduced to address the fairness issues of the original LC rates. We take these rates as given to showcase our counterfactual generation method. The choice of interest rates does not affect the development of our method.

\begin{figure}[h]
	\centering
	\caption{Interest Rate Distribution}
	\includegraphics[width=0.7\linewidth]{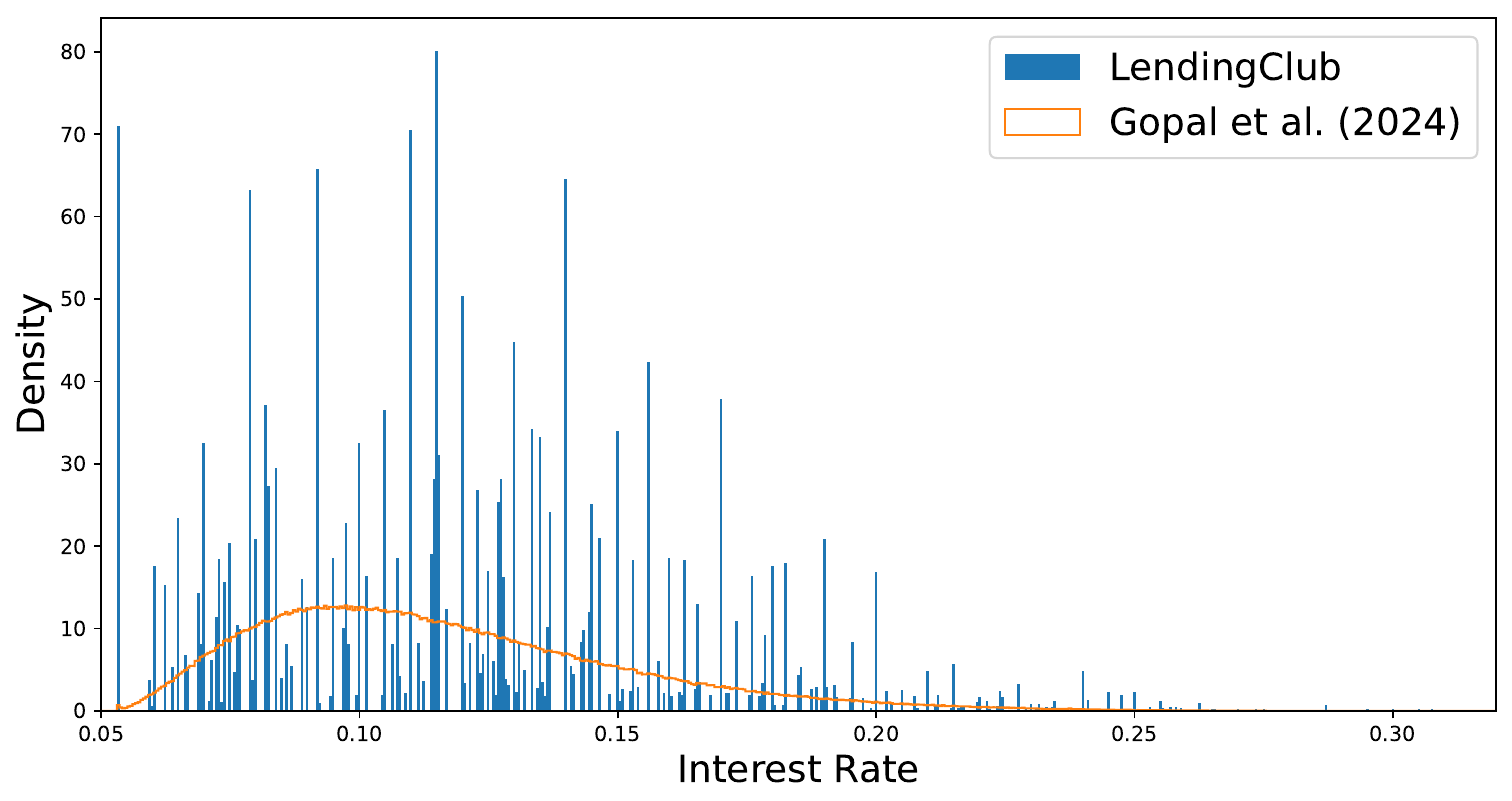}
	\label{fig:int_dist}
\end{figure}

Figure \ref{fig:int_dist} shows density plots of LendingClub rates and the alternative interest rates. The interest rates range from 5.3\% to 31.0\%. Both sets of rates have similar shapes and levels: They are right-skewed. The average interest rate under LendingClub is 11.9\%. The interest rates based on \cite{gopal2024pricing} are 0.3\% lower, with an average deviation of 2.8\% from the LC rates. The comparison of interest rates provides a baseline for the evaluation of counterfactuals.

\section{Loan Counterfactuals}\label{sec:cntfact}

To compare the LC rates against the interest rates based on \cite{gopal2024pricing} or other interest rate schemes, the counterfactual loan outcome is a crucial factor for lenders and platforms. 
Direct prediction of ROI is challenging. Consistent with the literature \citep{xia2021forecast}, we find that machine learning models, including neural networks and gradient-boosted regressions, cannot capture most of the variation in ROI, with out-of-sample R-squareds all smaller than 5\%. Therefore, we need to impose additional structure on the problem to improve its suitability for capturing counterfactual outcomes. We decompose the ROI prediction into the prediction of loan status and duration. Knowing when a loan is fully paid or charged off allows us to reconstruct cash flows and compute the ROI.

Since by definition counterfactuals are unobserved, it is infeasible to directly test the quality of counterfactuals. A model suitable for producing counterfactuals should, to some extent, accurately capture the underlying data-generating process of loan outcomes. This model should also be able to make precise predictions for actual loan outcomes. Therefore, strong predictive power of observed outcomes can be viewed as a necessary but not sufficient condition for the suitability of a model to generate counterfactuals. We first explore the predictive performance of GPT, and introduce counterfactual construction in Section \ref{sec:cntfact}.

\subsection{Prediction}\label{sec:pred}

As we directly observe loan outcomes under the original interest rates set by LendingClub, we can assess the predictive power of GPT for loan outcomes. We provide GPT with loan and borrower information, encoding any numerical information in text format, and ask GPT to predict whether and when the borrower would fully pay or be charged off. Based on this prediction, we compute the ROI of the loan. We compare the predictions from GPT with those from machine learning methods, adopting XGBoost for binary classification of loan status \citep{fu2021crowds} and gradient-boosted regression for ROI prediction \citep{xia2021forecast}. Since ML algorithms are designed to maximize predictive power, they serve as strong benchmarks. We randomly select 10,000 observations from the test set for our prediction exercise.\footnote{Because prediction is an intermediate step to producing counterfactuals and not the central focus of our study, we do not evaluate the predictive power of LLMs on the full test set to keep our costs down. Predicting 10,000 loans requires \$140 to run on GPT-3.5, in contrast to \$4,200 required for the full test set. We produce counterfactuals for our full test set.}

\subsubsection{Basic Prompt.}

We start with a simple prompt for the large language model: 
``I want you to act as a borrower with the following credit profile. Based on your credit profile and the loan specifics, predict your repayment."
Column (1) in Table \ref{tab:prompt} presents the results from this prompt. According to GPT, 1.7\% of loans will be charged off, which is considerably lower than the actual charged-off rate of 12.3\%. The LLM prediction yields an F1 score of 4.7\% and an AUC of 50.5\%, indicating limited capture of different loan statuses. Panel C includes the predictive performance for ROI. The R-squared of the LLM prediction is 1.97\%.\footnote{The raw predictions by the large language model yield a large negative out-of-sample R-squared. We conjecture that this may be due to an embedded bias in the language description within the pre-training corpus. To correct such a bias, we regress the actual ROI on the predicted ROI, $ROI = c_0 + c_1 \hat{ROI} + \epsilon$. Then, we form the bias-corrected prediction by $\tilde{ROI} = \hat{c}_0 + \hat{c}_1 \hat{ROI}$. The reported R-squared is based on the bias-corrected predictions. We perform the same procedure on ML predictions to ensure a fair comparison.}

The simple prompt for GPT does not lead to strong predictive power when compared to machine learning models. XGBoost achieves an F1 score of 35.4\% and an AUC of 64.3\%, while gradient-boosted regression has an R-squared of 3.48\%. Our initial LLM prompt performs worse than ML models across all three metrics. For a final evaluation, we borrow an idea from the econometrics literature, the forecast encompassing test \citep{chong1986econometric}. To evaluate two predictions for ROI, we run the following regression: 
\begin{equation}
    ROI_i = c + \varphi \hat{ROI}_{i,LLM} + (1-\varphi) \hat{ROI}_{i,ML} + \epsilon_i,
\end{equation}
where $\hat{ROI}_{i,LLM}$ and $\hat{ROI}_{i,ML}$ are the LLM and ML predictions for ROI, respectively, and $\epsilon_i$ is the error term. If one of the predictions is already optimal, it should carry a coefficient of one. The prediction that does not add incremental value will be assigned a coefficient of zero.

Coefficient combinations strictly between zero and one can be considered forecast combinations, in which the coefficient $\varphi$ captures how much the LLM prediction contributes to the combined prediction. If the coefficient is statistically significant, it indicates that the LLM identifies information meaningful to the prediction that is not captured by the machine learning algorithm. Otherwise, it suggests the prediction made by the machine learning algorithm encompasses that of the LLM. The higher the coefficient $\varphi$, the more informative the LLM prediction is compared to the ML prediction.

\begin{table}[h]
\centering
\caption{Prompt Engineering}\label{tab:prompt}
\begin{tabular}{l|ccccccc}
\hline \hline
                                                            & (1)                                                  & (2)                                                  & (3)                                                  & (4)                                                  & (5)                                                         & (6)                                                        & (7)                                                        \\ \hline
                                                            & \multicolumn{7}{c}{A: Prompt}                                                                                                                                                                                                                                                                                                                                                                                     \\ \hline
Role play                                                   & Borrower                                             & Expert                                               & Expert                                               & Expert                                               & Expert                                                      & 3 Experts                                                  & 4 Experts                                                  \\
Prompt type                                                 & \begin{tabular}[c]{@{}c@{}}Zero-\\ shot\end{tabular} & \begin{tabular}[c]{@{}c@{}}Zero-\\ shot\end{tabular} & \begin{tabular}[c]{@{}c@{}}Zero-\\ shot\end{tabular} & \begin{tabular}[c]{@{}c@{}}Zero-\\ shot\end{tabular} & \begin{tabular}[c]{@{}c@{}}Chain-of-\\ thought\end{tabular} & \begin{tabular}[c]{@{}c@{}}Tree-of-\\ thought\end{tabular} & \begin{tabular}[c]{@{}c@{}}Tree-of-\\ thought\end{tabular} \\
Information set                                             &                                                      &                                                      &                                                      &                                                      &                                                             &                                                            &                                                            \\
- Loan                                                      & \checkmark                                                    & \checkmark                                                    & \checkmark                                                    & \checkmark                                                    & \checkmark                                                           & \checkmark                                                          & \checkmark                                                          \\
- Borrower                                                  & \checkmark                                                    & \checkmark                                                    & \checkmark                                                    & \checkmark                                                    & \checkmark                                                           & \checkmark                                                          & \checkmark                                                          \\
- Platform                                                  &                                                      &                                                      & \checkmark                                                    & \checkmark                                                    & \checkmark                                                           & \checkmark                                                          & \checkmark                                                          \\
- ML                                                        &                                                      &                                                      &                                                      & \checkmark                                                    & \checkmark                                                           & \checkmark                                                          & \checkmark                                                          \\ \hline
                                                            & \multicolumn{7}{c}{B: Predictive Performance - Loan Status}                                                                                                                                                                                                                                                                                                                                                                     \\ \hline
Charged off                                                 & \multicolumn{1}{c}{1.7\%}                            & \multicolumn{1}{c}{36.1\%}                           & \multicolumn{1}{c}{27.6\%}                           & \multicolumn{1}{c}{20.0\%}                           & \multicolumn{1}{c}{30.1\%}                                  & \multicolumn{1}{c}{36.4\%}                                 & \multicolumn{1}{c}{36.3\%}                                 \\
F1 score                                                    & \multicolumn{1}{c}{4.7\%}                           & \multicolumn{1}{c}{25.9\%}                           & \multicolumn{1}{c}{25.9\%}                           & \multicolumn{1}{c}{29.7\%}                           & \multicolumn{1}{c}{33.8\%}                                  & \multicolumn{1}{c}{34.0\%}                                 & \multicolumn{1}{c}{34.0\%}                                 \\
AUC                                                         & \multicolumn{1}{c}{50.5\%}                           & \multicolumn{1}{c}{54.6\%}                           & \multicolumn{1}{c}{55.3\%}                           & \multicolumn{1}{c}{58.5\%}                           & \multicolumn{1}{c}{62.0\%}                                  & \multicolumn{1}{c}{62.7\%}                                 & \multicolumn{1}{c}{62.7\%}                                 \\ \hline
                                                            & \multicolumn{7}{c}{C: Predictive Performance - ROI}                                                                                                                                                                                                                                                                                                                                                                     \\ \hline
$R^2$                                                          & \multicolumn{1}{c}{1.97\%}                           & \multicolumn{1}{c}{1.97\%}                           & \multicolumn{1}{c}{1.96\%}                           & \multicolumn{1}{c}{2.31\%}                           & \multicolumn{1}{c}{2.76\%}                                  & \multicolumn{1}{c}{2.67\%}                                 & \multicolumn{1}{c}{2.84\%}                                 \\
\begin{tabular}[c]{@{}l@{}}Encompassing\\ test\end{tabular} & \multicolumn{1}{c}{1.71\%}                           & \multicolumn{1}{c}{-0.64\%}                          & \multicolumn{1}{c}{0.04\%}                           & \multicolumn{1}{c}{3.73\%***}                        & \multicolumn{1}{c}{4.64\%***}                               & \multicolumn{1}{c}{4.23\%***}                              & \multicolumn{1}{c}{5.19\%***}                              \\ \hline \hline
\end{tabular}
\end{table}

In our initial attempt of LLM-based loan prediction, the encompassing test shows a coefficient of 1.71\% on the LLM forecast $\hat{ROI}_{i,LLM}$. A positive coefficient suggests that the LLM does contain some incremental predictive power beyond that of the machine learning model, although it is not statistically distinguishable from zero. At first glance, GPT does not appear very useful in predicting loan outcomes or return on investment.

\subsubsection{Prompt Engineering.}

The performance of large language models (LLMs) can be significantly influenced by prompt engineering, which involves structuring instructions to maximize the desired outcomes. Emerging literature highlights that prompt engineering is crucial for complex tasks \citep{sahoo2024systematic}. In our study, we explore several techniques to enhance the performance of loan prediction, as shown in Table \ref{tab:prompt}. Columns (1) to (7) present increasingly sophisticated prompts and broader information sets, summarized in Panel A.

A different perspective leads to better prediction. Column (2) contains a prompt that provides the same information as Column (1), including loan and borrower characteristics, but with a change in the perspective of the response. We ask GPT to act as an expert in credit risk assessment rather than as a borrower. Column (3) further adds in platform-level loan statistics such as the average interest rate, overall charged-off rate, and average duration. This platform-level information describes an average loan on LendingClub, providing a benchmark for the LLM to compare a particular loan to all other loans. Columns (2) and (3) exhibit much higher predicted charged-off rates of 36.1\% and 27.6\%, respectively, along with more accurate classification results, as shown by the F1 score and AUC. The predictive performance remains unchanged compared to the initial prompt in Column (1). Specifically, the encompassing tests show that the LLM does not add significant value beyond the machine learning prediction.

Adding ML prediction to the LLM prompt further improves prediction. Column (4) adds to the prompt in Column (3) the ML prediction of loan status and duration, which leads to increases in the F1 score and AUC. The R-squared in the ROI prediction is 17\% larger compared to previous cases. The LLM demonstrates incremental predictive power for return on investment beyond that of the ML model with an encompassing test coefficient of 3.73\%, significant at the 1\% level.

\cite{kojima2022large} demonstrate that large language models can exhibit reasoning capabilities if the prompt directly asks for the thought process. We explore this ``chain-of-thought'' technique in Column (5). Classification metrics continue to improve compared to previous prompts, with the F1 score reaching 33.8\% and the AUC reaching 58.5\%. The R-squared also increases further to 2.76\%, a 19\% improvement compared to Column (4) without chain-of-thought. The encompassing test continues to show that the LLM holds additional predictive power compared to ML models.

\begin{figure}[h]
	\centering
	\caption{Loan Prediction Prompts and Responses}
	\includegraphics[width=0.9\linewidth]{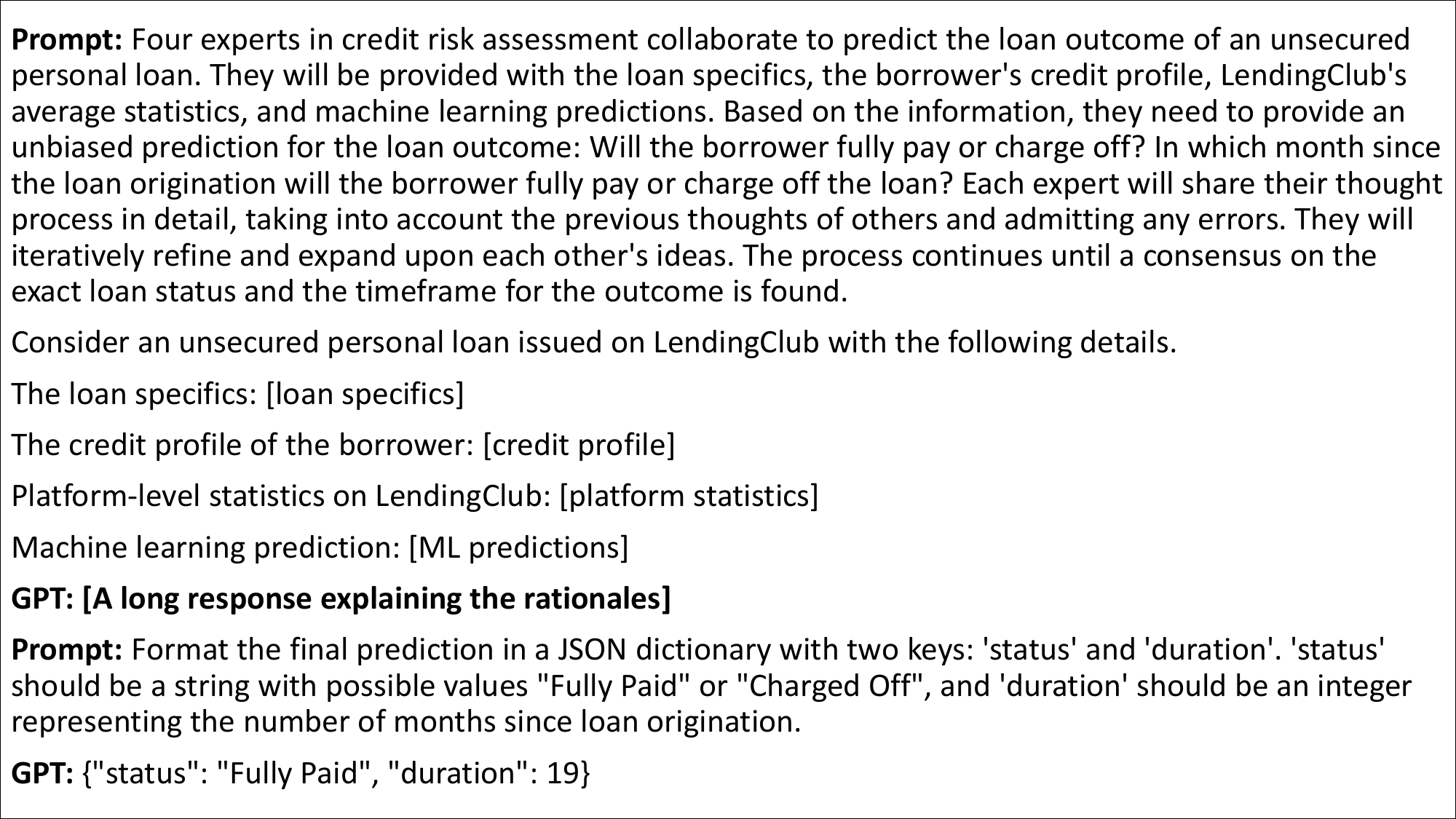}
	\label{fig:prompt_pred}
\end{figure}

Columns (6) and (7) use the ``tree-of-thought'' technique \citep{yao2024tree}. The LLM is asked to simulate a discussion among a panel of experts in credit risk assessment. The panel members are asked to share their thoughts, refine and expand upon one another, and discuss until a consensus response is reached. With three experts, the predictive performance on the classification problem of loan status and the regression problem of ROI is similar to that of the chain-of-thought prompt. With four experts, we further improve the performance on various metrics. The F1 and AUC scores are 34.0\% and 62.7\%, respectively, and the R-squared is 2.84\%. The predictive performance of the language model is now comparable to that of machine learning algorithms. The LLM coefficient in the encompassing test is 5.19\% and is statistically significant at the 1\% level.

Since the prompts in Columns (4) through (7) include ML predictions, improved LLM performance may simply result from regurgitating the ML information. If this were the case, the coefficient on the LLM prediction in the encompassing test would be zero, indicating that the LLM does not provide any additional predictive value beyond the ML model. However, the positive and statistically significant coefficients in encompassing tests strongly support that the LLM predictions carry extra predictive power beyond the ML models.

The finding that GPT can predict actual loan outcomes, with performance on par with machine learning models built for maximum predictive power, suggests that the LLM captures important aspects of borrower behavior. Figure \ref{fig:prompt_pred} presents the final prompt we use to generate the prediction (Column 7 in Table \ref{tab:prompt}). The LLM prediction is saved in JSON format for efficient ROI calculation.

\subsubsection{Discussion.}\label{sec:prompt_discuss}

The ROI prediction exercise leads to several interesting observations about large language models, which are potentially transferable to counterfactual analysis. First, consistent with \cite{shanahan2023role}, playing the appropriate role can significantly change predictive power. In prompt (1), we ask the large language model to act as the borrower provided in the prompt. In this case, only 1.7\% of the borrowers (LLM) report that their loans would be charged off, exhibiting a possible \textit{social desirability bias} \citep{grimm2010social} -- rather than giving truthful responses, individuals tend to respond to questions in a manner that they believe will be viewed favorably by others. As a result, the predictions given by the borrowers themselves are not informative about their actual repayment behavior. The F1 score of 4.7\% and the AUC of 50.5\% are similar to random guesses. The only change from prompt (1) to (2) is that we ask the language model to act as an expert on credit risk assessment and predict the loan outcome. Immediately, the predictions become more precise.\footnote{Role-playing can be as important as fine-tuning. We fine-tune the GPT-3.5 model with 665,185 samples by providing the language model with the question and answer pairs like in Figure \ref{fig:prompt_pred}, with the language model asked to act as a borrower. The AUC increases from 50.5\% to 54.0\% after fine-tuning. Although the computation cost is huge (a total of 0.4 billion tokens), the improvement in the predictive performance from fine-tuning with a poor prompt is much less than a simple change of perspective in the prompt. Prompt (4) has the same information set, but the AUC score is 58.5\% without fine-tuning.}

Second, combining a large language model with machine learning yields better results than using the LLM alone. A comparison of prompts (3) and (4) demonstrates that the inclusion of ML predictions significantly enhances the predictive performance of the language model. ML algorithms have a distinct advantage in handling structured data, while LLMs benefit from a more extensive information set and the logic reasoning ability derived from their pre-training corpus. Predictions based on the combination of LLM and ML leverage the strengths of both approaches.

The result that LLM $+$ ML $>$ LLM is consistent with the human-machine collaboration literature. \cite{lu2024information} illustrate that when extensive data and machine forecasts are present together, it can stimulate humans to actively reassess, thereby enhancing prediction accuracy. They reveal that humans are capable of naturally linking new features with previously ignored ones that could rectify the machine's errors. Table \ref{tab:prompt} shows similar results, indicating that LLMs inherit humans' cognitive advantages in complex tasks. Their findings provide a theoretical foundation for why LLM is a suitable candidate for counterfactual analysis and necessitate the inclusion of ML predictions in the prompt. 
The consistent results from our computer simulation and their field experiment suggest a potential method for conducting preliminary experiments by using LLMs to simulate humans before field experiments. This approach can offer insights into experiment design and reduce costs.

Lastly, chain-of-thought and tree-of-thought are effective prompting techniques to elicit logical reasoning in large language models. Their usage delivers a large boost in predictive power in our setting. The zero-shot prompt in prompts (1) to (4) asks the language model to directly output its prediction of loan status and duration without explaining how it makes such a prediction. 
The chain-of-thought prompting technique simply adds, ``Think through this logically and share your thought process."
Prompts (4) and (5) have the same information set and, thus, directly compare zero-shot and the chain-of-thought techniques. The result shows that chain-of-thought requires the large language model to form the prediction step by step, which improves the quality of the prediction. Tree-of-thought prompting takes logical reasoning one step further to have more than one voice in the discussion. This allows the language model to reflect on the predictions and make modifications. The results in Columns (6) and (7) show that this process of reflection and refinement further improves the prediction. Tree-of-thought prompt with four experts leads to the best predictive performance in our sample.

\subsection{Counterfactual Construction}\label{sec:cntfact}

The best model for prediction may not be the best model for constructing counterfactuals. Gradient-boosted regression has the highest R-squared in predicting ROI. However, the algorithm only models correlations among inputs and the prediction target. Changes in the interest rate of a loan can lead to changes in borrower behavior not observed in the training data, which may not be captured by a purely predictive model. As observed in the previous section, the large language model exhibits some ability to generate logically coherent reasoning for loan prediction. In this section, we create loan counterfactuals under the alternative interest rates using GPT.

\subsubsection{Methodology.}

Given the outcome of a loan with a particular interest rate, we ask the LLM what the outcome would be under a different interest rate. We follow the lessons learned from ROI prediction in constructing counterfactuals. First, we adopt tree-of-thought prompting and ask the LLM to act as a panel of four credit risk experts. Second, we include loan, borrower, and platform-level information in the prompt, and we also tell the LLM the actual loan outcome under the original interest rate set by LendingClub. This last piece of information serves as an anchor to link observed outcomes to unobserved counterfactuals. Third, we seek to combine the advantages of both machine learning models and LLMs by including in the prompt the counterfactual prediction made by the mixture cure model \citep{gopal2024pricing}.

\begin{figure}[h]
	\centering
	\caption{Counterfactual Construction based on LLM: Prompt}
	\includegraphics[width=0.9\linewidth]{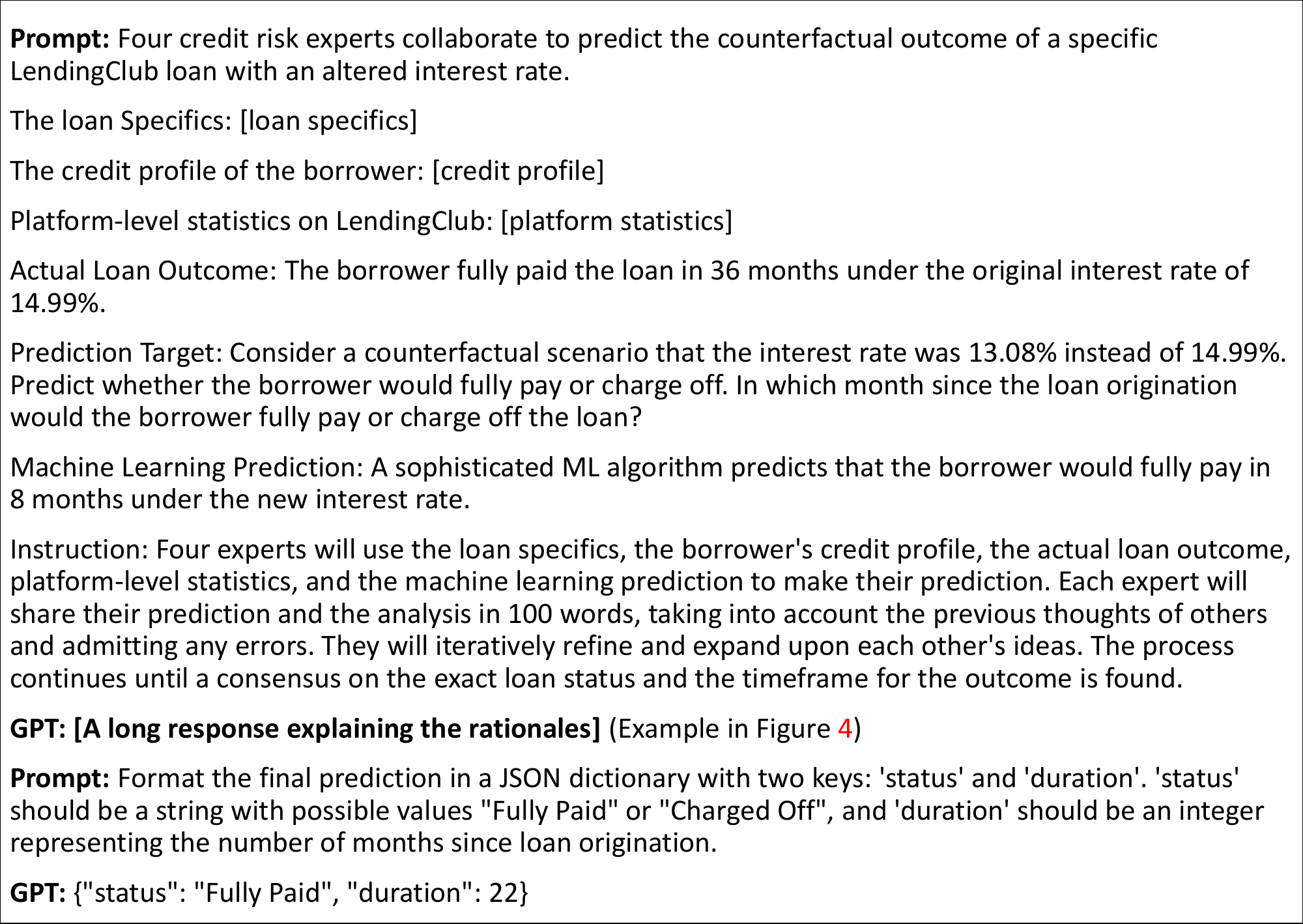}
	\label{fig:prompt_cntfact}
\end{figure}

Figure \ref{fig:prompt_cntfact} shows the prompt we use for generating the counterfactual for a particular loan with an interest rate of 14.99\% under LendingClub, which was fully paid in 36 months. 
Counterfactuals for other loans follow a similar prompt. 
We want to know the potential outcome if the interest rate were set to 13.08\%. The mixture cure model predicts that the borrower would pay off the loan in 8 months under the new interest rate. The LLM also predicts that the loan would be fully paid, but in 22 months.

\begin{figure}[h]
	\centering
	\caption{Counterfactual Construction based on LLM: Response (GPT-3.5)}
	\includegraphics[width=0.9\linewidth]{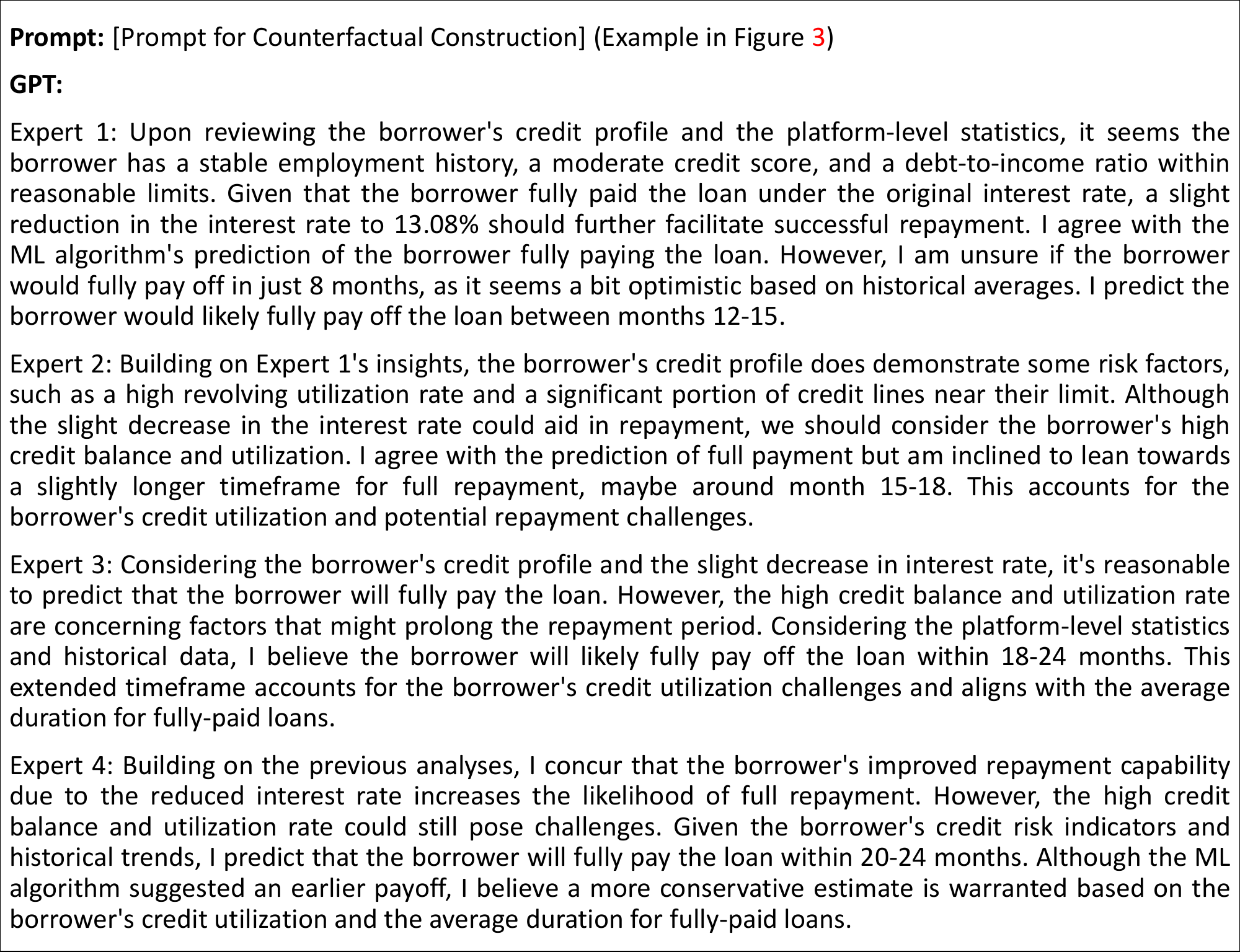}
	\label{fig:GPT_response}
\end{figure}

Figure \ref{fig:GPT_response} presents the detailed response from GPT. This response sheds light on several notable characteristics of large language models. The response from GPT attempts to explain the final prediction through what appears to be logically coherent reasoning. This process stands in contrast to prediction-oriented machine learning algorithms, which seem more opaque in comparison. The ability of the LLM to string together text into causal reasoning makes it well-suited for counterfactual analysis.

The LLM predicts a different outcome compared to the mixture cure model. Only 53.8\% of the LLM predictions on status and duration align with those of the mixture cure model. This result parallels loan prediction that although the prompt includes the ML prediction, the LLM prediction is not a mere repetition of it. Rather, the LLM critically evaluates the ML predictions based on its pre-trained knowledge and revises them with justifiable reason when necessary.

Large language models embed a degree of stochasticity that may be desirable when formulating counterfactuals. Predictions made by the LLM are not deterministic; two identical prompts may receive two different responses. This stochastic nature of LLMs reflects the inherent uncertainty and variability in borrower repayment behavior and loan outcomes.

\subsubsection{Counterfactual Loan Outcomes.}

We compare our LLM-based counterfactual method with the mixture cure model presented in \cite{gopal2024pricing}. Both methods are applied to generate counterfactual loan outcomes for the loans in the test sample under the alternative interest rates. Table \ref{tab:outcome} presents summary statistics for these counterfactual outcomes and compares them to the actual outcomes under the original LC rates. The average interest rate under LendingClub is 11.8\% with a standard deviation of 3.9\%. The average interest rate under GCPP is 0.2\% lower than that of LendingClub.

\begin{table}[h]
	\centering
	\caption{Summary of Counterfactual Loan Outcomes}\label{tab:outcome}
    \begin{tabular}{l|cc|crrrr}
    \hline \hline
                          & \multicolumn{2}{c|}{Interest Rate}              & \multicolumn{5}{c}{Outcome}                                                                                                                                                                                                                                                                \\ \hline
    \textbf{}             & Mean                     & SD                      & \multicolumn{1}{c|}{\begin{tabular}[c]{@{}c@{}}Counter-\\ factual\end{tabular}} & \multicolumn{1}{c}{ROI} & \multicolumn{1}{c}{SD(ROI)} & \multicolumn{1}{c}{Utility} & \multicolumn{1}{c}{\begin{tabular}[c]{@{}c@{}}Charged-\\ off rate\end{tabular}} \\ \hline
    LendingClub                    & 11.8\%                  & 3.9\%                  & \multicolumn{1}{c|}{NA}                                                         & 4.36\%                  & 24.08\%                                                                & -1.1880                     & 12.34\%                                                                         \\ \hline
    \multirow{2}{*}{\cite{gopal2024pricing}} & \multirow{2}{*}{11.6\%} & \multirow{2}{*}{3.3\%} & \multicolumn{1}{c|}{Survival}                                                   & 4.34\%                  & 10.09\%                                                                & -0.9931                     & 6.22\%                                                                          \\
                          &                          &                         & \multicolumn{1}{c|}{LLM}                                                        & 3.68\%                  & 17.01\%                                                                & -1.0396                     & 9.65\%                                                                          \\ \hline\hline
    \end{tabular}
\end{table}

Under LC rates, the average return on investment is 4.36\%, with a standard deviation of 24.08\%. According to both counterfactual construction methods, the ROI under the alternative interest rates is lower compared to LC rates (4.34\% from the mixture cure model and 3.68\% from the LLM), which is consistent with the lowered interest rates. At the same time, the counterfactual ROIs are less variable, with standard deviations considerably lower than the actual ROIs.

The last column shows that 12.34\% of loans on LendingClub are charged off. The counterfactual outcome from the mixture cure model indicates that the charged-off rate, when borrowers are faced with the alternative interest rates, is just 6.22\%. 
Such a nearly 50\% decrease in the charged-off fraction seems rather high. According to the LLM counterfactuals, 9.65\% of loans are charged off, closer to LendingClub's value. This observation highlights that GPT can correct the potential mistakes made by the ML algorithm.

Although it is difficult to assess if a counterfactual outcome is accurate, two logical principles specific to our context provide some guidance on whether the prediction is sensible: 1) A charged-off loan remains charged-off if the interest rate was increased, and 2) A fully-paid loan remains fully paid if the interest rate was decreased. These intuitive rules are encoded in the design of the survival approach. We analyze all counterfactuals generated by the LLM and find that 85.3\% abide by rule 1) and 99.6\% abide by rule 2).\footnote{We also tried the latest \textit{GPT-4o} model (released in May 2024) on 10,000 random samples and found that 100\% of the predictions follow the two rules. It seems that the reasoning capability of \textit{GPT-4o} exceeds that of \textit{GPT-3.5}.} These results indicate that the vast majority of the LLM predictions follow a certain logical reasoning appropriate in our setting.

\section{Conclusion}\label{sec:conclusion}

This paper demonstrates that large language models are a suitable approach for counterfactual analysis within the context of online lending. Our exploration yields two important takeaways that could be applicable to other contexts. First, we find that prompt engineering is as crucial as fine-tuning. Specifically, techniques such as role-playing, chain-of-thought, and tree-of-thought are effective in eliciting logical reasoning. Second, we observe that large language models exhibit several interesting human-like behaviors, including social desirability bias and incremental value addition to machine learning algorithms in complex tasks. The human-mimicking behavior of large language models suggests the potential for using LLMs to simulate humans in field experiments and surveys as a preliminary test, providing insights into experiment design and reducing costs.

\bibliography{pricing.bib}

\end{document}